\pdfoutput=1

\documentclass[final,5p,times,twocolumn]{elsarticle}

\usepackage{graphicx}
\usepackage{amssymb}

\usepackage{lineno}
\usepackage{booktabs}

\usepackage{algorithm2e}
\usepackage{subfig}
\usepackage{float}
\usepackage{nameref,zref-xr} 
\zxrsetup{toltxlabel} 
\zexternaldocument*[supp-]{supp} 

\usepackage{hyperref}

\journal{Arxiv}

\begin{document}

\begin{frontmatter}



\title{Uncovering convolutional neural network decisions for diagnosing multiple sclerosis on conventional MRI using layer-wise relevance propagation}



\author[label1,label2]{Fabian Eitel}
\author[label1,label2]{Emily Soehler}
\author[label4,label5]{Judith Bellmann-Strobl}
\author[label3,label4,label8]{Alexander U. Brandt}
\author[label3]{Klemens Ruprecht}
\author[label3,label4]{Ren\'e M. Giess}
\author[label3,label4,label5]{Joseph Kuchling}
\author[label3,label4,label5]{Susanna Asseyer}
\author[label3,label4]{Martin Weygandt}
\author[label2,label7]{John-Dylan Haynes}
\author[label3,label4,label6]{Michael Scheel}
\author[label3,label4,label5,label7]{Friedemann Paul\fnref{s1}}
\author[label1,label2]{Kerstin Ritter\corref{cor1}\fnref{s1}}

\address[label1]{Charit\'e -- Universit\"atsmedizin Berlin, corporate member of Freie
Universit\"at Berlin, Humboldt-Universit\"at zu Berlin, and Berlin Institute of
Health (BIH); Department of Psychiatry and Psychotherapy; 10117 Berlin, Germany.}

\address[label2]{Charit\'e -- Universit\"atsmedizin Berlin, corporate member of Freie
Universit\"at Berlin, Humboldt-Universit\"at zu Berlin, and Berlin Institute of
Health (BIH); Berlin Center for Advanced Neuroimaging, Bernstein Center for Computational Neuroscience; 10117 Berlin, Germany.}

\address[label3]{Charit\'e -- Universit\"atsmedizin Berlin, corporate member of Freie
Universit\"at Berlin, Humboldt-Universit\"at zu Berlin, and Berlin Institute of
Health (BIH); Department of Neurology; 10117 Berlin, Germany.}

\address[label4]{Charit\'e -- Universit\"atsmedizin Berlin, corporate member of Freie
Universit\"at Berlin, Humboldt-Universit\"at zu Berlin, and Berlin Institute of
Health (BIH); NeuroCure Clinical Research Center; 10117 Berlin, Germany.}

\address[label5]{Charit\'e -- Universit\"atsmedizin Berlin, corporate member of Freie Universit\"at Berlin, Humboldt-Universität zu Berlin, and Berlin Institute of Health (BIH); Experimental and Clinical Research Center, Max Delbr\"uck Center for Molecular Medicine; 10117 Berlin, Germany.}

\address[label6]{Charit\'e -- Universit\"atsmedizin Berlin, corporate member of Freie Universit\"at Berlin, Humboldt-Universität zu Berlin, and Berlin Institute of Health (BIH); Department of Neuroradiology; 10117 Berlin, Germany.}

\address[label7]{Einstein Center for Digital Future Berlin, Germany.}

\address[label8]{Department of Neurology, University of California, Irvine, California, USA.}
\cortext[cor1]{Corresponding author: kerstin.ritter@charite.de}
\fntext[s1]{Shared authorship}

\begin{abstract}
Machine learning-based imaging diagnostics has recently reached or even superseded the level of clinical experts in several clinical domains. However, classification decisions of a trained machine learning system are typically non-transparent, a major hindrance for clinical integration, error tracking or knowledge discovery.
In this study, we present a transparent deep learning framework relying on convolutional neural networks (CNNs) and layer-wise relevance propagation (LRP) for diagnosing multiple sclerosis (MS), the most widespread autoimmune neuroinflammatory disease. MS is commonly diagnosed utilizing a combination of clinical presentation and conventional magnetic resonance imaging (MRI), specifically the occurrence
and presentation of white matter lesions in T2-weighted images. 
We hypothesized that using LRP in a naive predictive model would enable us to uncover relevant image features that a trained CNN uses for decision-making. Since imaging markers in MS are well-established this would enable us to validate the respective CNN model.  
First, we pre-trained a CNN on MRI data from the Alzheimer's Disease Neuroimaging Initiative ($n = 921$), afterwards specializing the CNN to discriminate between MS patients and healthy controls ($n=147$). 
Using LRP, we then produced a heatmap for each subject in the holdout set depicting the voxel-wise relevance for a particular classification decision. The resulting CNN model resulted in a balanced accuracy of 87.04\% and an area under the curve of 96.08\% in a receiver operating characteristic curve. The subsequent LRP visualization revealed that the CNN model focuses indeed on individual lesions, but also incorporates additional information such as lesion location, non-lesional white matter or gray matter areas such as the thalamus, which are established conventional and advanced MRI markers in MS.
We conclude that LRP and the proposed framework have the capability to make diagnostic decisions of CNN models transparent, which could serve to justify classification decisions for clinical review, verify diagnosis-relevant features and potentially gather new disease knowledge.

\end{abstract}


\end{frontmatter}


\section{Introduction}
\label{S:intro}
Multiple Sclerosis (MS) is the most widespread autoimmune neuroinflammatory disease in young adults with 2.2 million cases reported worldwide \citep{GBD2016MultipleSclerosisCollaborators2019}. The disease is mainly characterized by inflammation, demyelination and neurodegeneration in the central nervous system and often leads to substantial disability in patients \citep{Reich2018}. The current quasi-standard for diagnosing MS, the McDonald criteria, relies on clinical presentation and the presence of lesions visible in conventional T2-weighted brain magnetic resonance imaging (MRI) data \citep{Thompson2018}. Most common are fluid-suppressed T2-weighted image sequences, which are sensitive towards MS-relevant white matter lesions, but also relatively unspecific with respect to underlying disease processes \citep{Geraldes2018}. Several other imaging markers have been described including neurodegeneration, thalamic atrophy, cortical lesions, altered functional connectivity or central vein signs \citep{Lowe2002, Azevedo2018, Absinta2016, Filippi2016, Sati2016, Backner2018}, of which some are captured in conventional MRI and others require advanced MRI techniques such as diffusion weighted imaging or functional MRI. 

In the last decade, a lot of research effort has been put on the automatic (i.e. data-driven) detection of neurological diseases based on neuroimaging data including MRI \citep{Orru2012, Woo2017a}. Early approaches combined parameter-based machine learning algorithms, such as support vector machines, with carefully extracted features known or hypothesized to be relevant in the respective disease. In MS research, features ranging from T2 lesion characteristics to local intensity patterns or multi-scale information extracted from MRI data have been used in combination with standard machine learning analyses to either diagnose MS or predict disease progression \citep{Weygandt2011, Hackmack2012a, Hackmack2012b, Weygandt2015, Wottschel2015}. 
While choosing features based on expert criteria reflects the current state of knowledge, it does not allow for finding new and potentially unexpected hidden data properties, which might also help in characterizing a certain disease. Deep learning techniques fill a gap here and allow for utilizing hierarchical information directly from raw or minimally processed data \citep{LeCun2015}. By being specifically tailored to image data, in particular convolutional neural networks (CNNs) have led to major breakthroughs in medical imaging \citep{Litjens2017, Rajpurkar2017a,Rajpurkar2017b,DeFauw2018ClinicallyDisease}. In neuroimaging, most CNN analyses so far focused on Alzheimer's disease \citep{Vieira2017}, but there are also some recent studies in MS. Given the importance of lesions in diagnosing MS and monitoring disease progression, most efforts have been put on the task of lesion segmentation \citep{Valverde2017, Brosch2016, Khastavaneh2017}. 
Others used CNNs to diagnose MS based on 2-dimensional MRI slices \citep{Wang2018} or to predict short-term disease activity based on binary lesion masks \citep{Yoo2016b}. 

Despite their potential, deep learning methods are criticized for being non-transparent (such as a `black box') due to the difficulty to retrace the classification decision in light of huge parameter spaces and highly non-linear interactions \citep{Castelvecchi2016}. This is especially problematic in medical applications since understanding and explaining neural network decisions is required for clinical integration, error tracking or knowledge discovery.
Explaining neural network decisions is an open research area in computer science and a number of suggestions have been made in recent years. 
Different directions for explanations include visualizing features \citep{Zeiler2014}, generating images that maximally activate a certain neuron \citep{olah2017feature} and creating heatmaps based on the input images indicating the relevance of each voxel for the final classification decision \citep{ Simonyan2014,Bach2015, Springenberg2015}.
Heatmaps are in particular valuable in the medical context, since they allow for an easy and intuitive investigation of what the respective classifier found to be important directly in the input data. 
Besides understanding diagnostic decisions for individual patients, heatmaps might be useful in validating CNN models. 
Recently, we have shown the potential of transparent CNN applications for knowledge discovery in Alzheimer's disease \citep{rieke2018visualizing, boehle2019Visualizing}. 

The objective of the current study was to investigate whether a transparency approach can uncover decision processes in MRI-based diagnosis of MS, a disease with well-defined imaging markers, thereby supporting future clinical implementation and verification of machine learning-based diagnosis systems. We present a transparent CNN framework for the MRI-based diagnosis of clinically definite MS relying on layer-wise relevance propagation (LRP, \citep{Bach2015, Samek2015}) -- a heatmap method that has been shown to outperform previous approaches in terms of explainability and disease-specific evidence \citep{Samek2015,boehle2019Visualizing}. Since the data set was rather small ($n=147$), we investigated the effect of pre-training the CNN on data from the Alzheimer's Disease Neuroimaging Initiative (ADNI, $n = 921$). Using LRP, invididual heatmaps were generated for each subject and analyzed with respect to well-established imaging features in MS (e.g. white matter lesions or thalamus atrophy). 
By showing that LRP in combination with a naive CNN model (i.e. a model independent of MS-specific knowledge) indeed helps in uncovering relevant imaging features, we conclude that this framework is not only useful in justifying individual diagnostic decisions but also to validate CNN models (especially in light of small sample sizes).

\begin{figure*}
    \centering
    \includegraphics[width=0.8\textwidth]{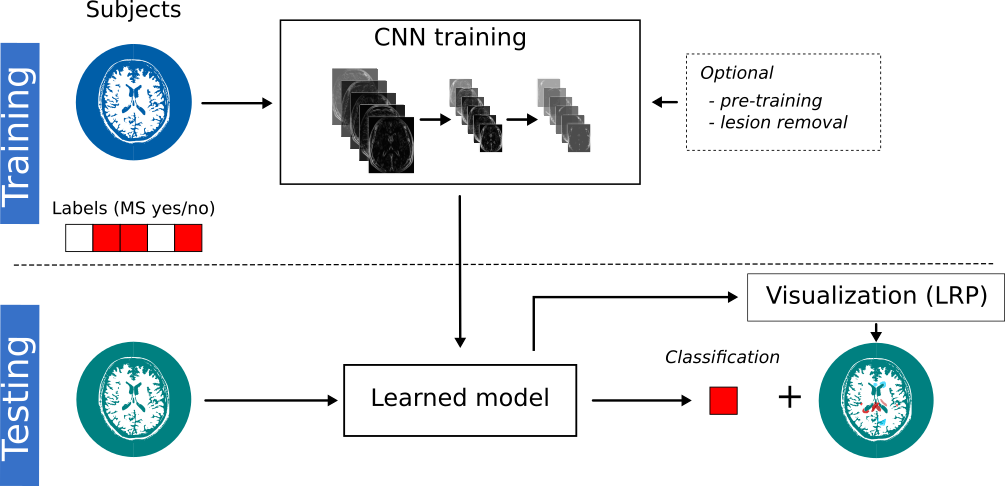}
    \caption{Illustration of the transparent CNN framework. In the training phase, the CNN model learns a non-linear relationship between the MRI data and the binary diagnostic labels (MS yes/no). Optionally, the CNN models are pre-trained on a substitute data set or lesions are filled in the MRI data. The learned CNN model is then tested on new subjects to predict the diagnostic label. By supplementing this label with a LRP heatmap, which indicates the relevance of each voxel for the respective label, this framework allows us to understand (at least to some extent) the classification decision in individual subjects. Additionally, the validity of the CNN models can be assessed by matching highlighted brain areas with domain knowledge.
    }
    \label{fig:overview}
\end{figure*}

\section{Materials and methods} 
\label{S:methods}

\subsection{Subjects}
In the present study, we retrospectively analyzed data collected by FP from Charit\'e -- Universit\"atsmedizin Berlin as part of the VIMS study: Follow-up examination of visual parameters for the creation of a database (neuro-ophthalmologic register) in patients with MS versus healthy subjects.\footnote{\url{https://neurocure.de/en/clinical-center/clinical-studies/current-studies.html}} 
We enrolled 76 patients with clinically definite multiple sclerosis (MS) according to the McDonald criteria 2010 \citep{Polman2011} and 71 healthy controls. Patients were excluded if they were outside the age range of 18 -- 69 or did not have an MRI scan. 
All patients were examined under supervision of a board-certified neurologist at the NeuroCure Clinical Research Center (Charit\'e -- Universit\"atsmedizin Berlin) between January 2011 and July 2015.
All participants provided written informed consent prior to their inclusion in the study. The study was approved by the local ethics committee and was performed in accordance with the 1964 Declaration of Helsinki in its currently applicable version.
Part of this data has been used in previous studies (e.g. \citep{Kuchling2018}).
Demographical details of subjects can be found in Table \ref{tab:demo}.

\begin{table*}[h]
\centering\small
\begin{tabular}{l c c}
\hline
& \textbf{MS patients} & \textbf{Healthy controls}\\
\hline
\textbf{Subjects} [n] & 76 &  71\\
\textbf{Female/Male}, in \% &  55 \% / 45 \% &  65 \% / 35 \% \\
\textbf{Age}, mean $\pm $ std &  43.32 ($\pm $ 11.99) & 38.23 ($\pm $ 13.10) \\
\textbf{Disease duration}, mean $\pm $ std & 149.65 ($\pm $ 123.35) & n.a. \\
\textbf{EDSS}, median, range & 2.50 (0.00 - 6.50) & n.a. \\
\textbf{Lesion volume}, mean $\pm $ std & 7.28 ($\pm $ 8.09) & 0.57 ($\pm $ 1.94)\\
\hline
\end{tabular}
\caption{Demographics of MS patients and healthy controls. Disease duration is measured in months and lesion volume in ml. EDSS, expanded disability status scale; std, standard deviation.}
\label{tab:demo}
\end{table*}

\subsection{MRI acquisition and preprocessing}
All MRI data were acquired on the same 3 T scanner (Tim Trio Siemens, Erlangen, Germany) using a volumetric high-resolution T1 weighted magnetization prepared rapid acquisition gradient echo (MPRAGE) sequence (TR = 1900 ms, TE = 2.55 ms, TI = 900 ms, FOV=240x240 mm$^2$, matrix 240x240, 176 slices, slice thickness 1 mm) as well as a volumetric high-resolution fluid-attenuated inversion recovery sequence (FLAIR, TR = 6000 ms, TE = 388 ms, TI = 2100 ms; FOV=256x256 mm$^2$, slice thickness 1 mm). Individual lesion masks were generated based on FLAIR images by three expert raters under the supervision of a board-certified radiologist using ITK-SNAP\footnote{\url{www.itksnap.org}} \citep{Yushkevich2006}. The MRI data were preprocessed using a customized pipeline based on the software programs statistical parametric mapping (SPM) \citep{spm_book}, Advanced Normalization Tools (ANTs) \citep{Avants2011} and FMRIB Software Library (FSL) \citep{smith2004advances}. After a bias-field correction and field of view-cropping, MPRAGE images were linearly registered to the Montreal Neurological Institute (MNI)-template. FLAIR images were coregistered to the native MPRAGE images and then transformed to the MNI space using the transformation matrix estimated on the the MPRAGE images. Normalization parameters were estimated based on MPRAGE images because they provide a better tissue contrast than FLAIR. For each person, the skull was extracted using the Brain Extraction Tool (BET) of FSL. Please note that only FLAIR data entered the subsequent analyses and that they were preprocessed in that way to ensure that images are in relative realignment while preserving individual structural variations. 
For computational efficiency initial scan volumes (182x218x182) were down-sampled to 96x114x96 voxels and standardized for each subject using min-max scaling. 
To analyze the impact of white matter lesions, we generated an additional MRI data set, in which the lesions were filled. For this, we implemented a version of \citep{Valverde2014}, in which lesion areas (according to the manually segmented lesion masks) have been replaced by local average intensities in normal-appearing white matter. White matter maps were obtained from the SPM 12 tissue segmentation algorithm \citep{AshburnerSegmentationSPM}.

\subsection{ADNI data for pre-training}
Data used for pre-training were obtained from the Alzheimer’s Disease Neuroimaging Initiative (ADNI) database\footnote{\url{http://adni.loni.usc.edu}, RRID:SCR\_003007}. We have used subjects from ADNI phase 1 who were included in one of two standard MRI collections \citep{Wyman2013}. We only selected MRI data of Alzheimer's disease patients and cognitive normal subjects, in total 921 MRI scans from 276 subjects (covering one to three time points). The MRI scans were acquired with 1.5 Tesla scanners at multiple sites and had already undergone gradient non-linearity, intensity inhomogeneity and phantom-based distortion correction. T1-weighted MPRAGE scans were downloaded and warped to MNI space with ANTs \citep{Avants2011}. As for the MS data, the initial scan volumes were down-sampled to 96x114x96 voxels and standardized.

\subsection{Classification and visualization analyses}
Based on the preprocessed FLAIR data, we first trained several convolutional neural network (CNN) models (with and without pre-training, with and without lesion-filling) to discriminate MS patients and healthy controls and then explained the model's decisions for individual subjects in the test data using layer-wise relevance propagation (LRP). For the CNN models, we evaluated the effect of transfer learning by (1) training the model solely on MS data and (2) pre-training the model on ADNI data and fine-tuning it on MS data. 
To examine whether our pre-trained network can also learn from only normal-appearing brain matter (NABM), i.e. regions without hyper-intense lesions, we retrained the network on lesion-filled MS data. 
As baseline analyses, we included a support vector machine to classify based on (1) lesion volume and (2) preprocessed FLAIR data.  
Prior to training, the MS data set was randomly split into two sets: (1) a set for training and hyperparameter optimization (85 \%) and (2) a holdout set used only for final model evaluation (15 \%). The code for all models and also the lesion filling algorithm is available at \url{https://github.com/derEitel/explainableMS}.
In the following subsections, we specify our parameter settings for CNNs, transfer learning and visualization techniques (in particular LRP).

\subsubsection{Convolutional neural networks}
In this study, we used a convolutional neural network (CNN) architecture consisting of four convolutional layers followed by exponential linear units (ELUs) activation functions and four max-pooling layers applied after the first, second and fourth ELU activation. For each convolutional layer, we learned 64 filters with a kernel size of 3x3x3. Finally, a linear layer with an output shape of 1 and a sigmoid activation returns the classification score. 
The rationale behind this architecture was mainly to avoid overfitting and therefore has a comparably low number of trainable parameters (namely 333,889). To improve generalization further, the model has been regularized using a dropout on the outputs of each max-pooling layer (\(p = 0.3\)), L2-regularization (\(\lambda = 0.01\)) using the weights of the third and fourth convolutional layer, and finally early-stopping the training after the validation loss has not improved for 15 epochs. Additionally, the data was augmented during training by flipping it along the sagittal axis with a probability of 50\% and randomly translated between -2 and 2 pixels along the sagittal axis. We trained all models using the Adam optimizer \citep{Kingma2014}. 
Hyperparameters were optimized on 85\% of the training data, leaving 15\% for validation. After finding suitable hyperparameters, the model performance was tested out-of-sample on the holdout set. All CNN experiments were repeated 10 times, and thus reported metrics are an average over all 10 trials. We report balanced accuracy as a mean between sensitivity and specificity as well as area under the receiver operating characteristic curve (AUC). All code was implemented using Keras \citep{chollet2015keras} with the TensorFlow \citep{tensorflow2015-whitepaper} backend.\footnote{Keras version 2.2; TensorFlow version 1.11}

\subsubsection{Transfer learning}
Due to the small sample size of the MS data set, we employed the principle of transfer learning \citep{Crammer2008, Duan2009, Ben-David2010}, which has been shown to improve performance in medical imaging including MRI data \citep{Gupta2013, tajbakhsh2016convolutional, Ghafoorian2017TransferLearning, Hosseini-Asl2018, Basaia2019}. 
We pre-trained our CNN model on ADNI data to separate Alzheimer patients and healthy controls, and fine-tuned it on the MS data set to separate MS patients and healthy controls. 
Since the ADNI data set contains multiple scans for several subjects we ensured that validation and testing was done on disjoint subject sets. The average balanced accuracy over all trials was 78.47 \%. For further analysis, we selected a model from the 10 trials based on its performance, and then picked its training checkpoint with the best validation accuracy of 82.50 \%. Fine-tuning on the MS data set uses the same model architecture, which is initialized with the weights and biases of the selected pre-trained model instead of randomly distributed values. 
Please note that we here transferred a CNN model (1) across diseases (Alzheimer's disease to MS) and (2) across MRI sequences (MPRAGE to FLAIR).

\subsubsection{Visualization}
Deep learning methods are often criticized for their lack of interpretability \citep{Castelvecchi2016, Montavon2018, Lapuschkin2019}. In contrast to the decision nodes of a decision tree, which can be easily followed, and standardized coefficients in regression analysis which can determine feature importance, the learned weights of a CNN and its non-linear combinations are harder to comprehend. Over the last years much research has focused on improving the interpretability of neural networks. While some work has focused on understanding class representations and functions of individual neurons, others have developed methods to generate heatmaps based on the input data that indicate the importance or relevance of each pixel or voxel for the final classification decision \citep{Bach2015,Simonyan2013, Springenberg2015}. The latter approach is in particular promising in the medical field since it allows for explaining in a fast and intuitive way individual classification decisions without the need for delving deeply into the network structure \citep{boehle2019Visualizing}. However, heatmaps can be computed in different ways and it is important to understand their underlying principles. 
 The most popular approach is sensitivity analysis \citep{Simonyan2013}, in which the norm $\bigg\| \cdot \bigg\|_{l_{p}}$ over the gradient of the classification score $f(x)$ with respect to each pixel or voxel $x_i$ is calculated:
  \begin{equation}\label{eq:sensitivity_analysis}
R_i = \bigg\| \frac{\partial}{\partial x_i} f(x) \bigg\|_{l_{p}}    
\end{equation}
 
 This results in image-specific saliency maps attributing higher scores to voxels, for which small variations largely affect the classification score. There exist different variations and modifications of sensitivity analysis (e.g. \citep{Springenberg2015}). However, sensitivity analyses and its variants are very susceptible to highlight regions which would change the classification score drastically if altered but might not be relevant to a human expert. Examples include pixels in which the main object is hidden by another object or possible artifacts in the data set. Furthermore sensitivity analysis highlights regions regardless of their importance for or against a class but for classification in general. When applied in neuroimaging this could cause the visualization to emphasize brain regions which have a general relevance for the given task (e.g. hippocampus in classifying Alzheimer's disease patients) but not an individual directed relevance for the specific subject \citep{boehle2019Visualizing}. 
 
 Recently, layer-wise relevance propagation (LRP) has been introduced as an alternative method for producing heatmaps and shown to be superior to sensitivity analysis \citep{Bach2015, Samek2015, Lapuschkin2019}. 
 LRP uses the classification score \(f(x)\) directly (and not the gradient) and propagates it through the network using the following rule

\begin{equation}\label{eq:LRP}
    R_i = \sum_j \frac{x_i w_{ij}}{\sum_i x_i w_{ij} + \epsilon} R_{j}.
\end{equation}
Here, the relevance from layer \(R_j\) is propagated to its previous layer \(R_i\). The term \(\epsilon\) can be set to a small value (in this study: 0.001) to avoid division by 0. By using both the activation \(x_i\) as well as the weights \(w_{ij}\) connecting layers \(i\) and \(j\), LRP assigns a larger share to neurons that are more strongly activated and to connections which have been reinforced during training \citep{samek2017explainable}. By decomposing the classification score \(f(x)\) rather than the gradient and conserving the classification score during backpropagation, LRP overcomes the flaws of sensitivity analysis \cite{samek2017explainable} and has been shown to provide evidence for Alzheimer's disease in individual subjects  \citep{boehle2019Visualizing}. 

In this study, we produced individual LRP heatmaps for every subject in the holdout set. We have used the iNNvestigate implementation of LRP \citep{INNvestigateLRPAlber}.\footnote{The implementation can be found at \url{https://github.com/albermax/innvestigate}} Besides qualitatively comparing individual heatmaps, we compared average heatmaps of MS patients and healthy controls. We evaluated the importance of different brain regions by computing the average relevance for each brain area in the (1) Neuromorphometrics atlas\footnote{Contained in the SPM12 software, \url{https://www.fil.ion.ucl.ac.uk/spm/software/spm12/}} \citep{Bakker2015} and the (2) JHU DTI-based white-matter atlas\footnote{\url{https://fsl.fmrib.ox.ac.uk/fsl/fslwiki/Atlases}} \citep{Mori2005}. To evaluate the effect of transfer learning on the heatmaps, we compare average heatmaps for MS patients before and after pre-training. To assess the relevance of normal-appearing brain areas in contrast to lesion areas, we computed relevance scores separately for the original MRI data set and the lesion filled MRI data set.

\subsubsection{Baseline analyses}
As a baseline we have trained a support vector machine (SVM) to classify between MS patients and healthy controls based on T2 lesion load. Hyperparameters were tuned on the training data set using grid search, nested within a 5-fold cross-validation (SVM kernel: linear and radial basis function [RBF], $C, \gamma=[0.001, 0.1, 1, 10]$). For completeness we have also trained a SVM on the preprocessed FLAIR images with the same cross-validation strategy, optionally together with a prior dimensionality reduction via principal component analysis.

\section{Results}
\label{S:results}

\subsection{Classification performance}
In Table \ref{res-table}, we depict the performance for the different classification models. As expected T2 lesion load -- as one of the core biomarkers in MS -- in combination with a SVM led to a high balanced accuracy (88.46\%) and a high area under the curve (AUC) of the receiver operating characteristic (94.62\%). When instead of the T2 lesion load the entire FLAIR image is used as input to the SVM, the AUC dropped down to 66.92\%. The CNN model solely trained on the MS data set resulted in a balanced accuracy of 71.23\% and an AUC of 85.46\%. When the network has been pre-trained on the ADNI data set and fine-tuned to the MS data set, the balanced accuracy increased by 16 percentage points to 87.04\% and is therefore comparable to the performance of the baseline T2 lesion load model. Moreover, the pre-trained CNN model outperformed all other classifiers in terms of AUC (96.08\%) and importantly also in terms of sensitivity (93.08\%). The ROC curve for all 10 trials is shown in supplementary Figure 1.
To assess the impact of normal-appearing brain matter, we trained the same CNN model on lesion-filled FLAIR data. Still, a reasonable balanced accuracy of 70.15 \% and a relatively high AUC of 90.92 \% has been achieved.

 \begin{table*}
   \footnotesize
   \centering
   \begin{tabular}{cccccccc}
     \toprule
     Data & Pre-train. & Class.     & Bal. acc.     & Sens. & Spec. & AUC \\
     \midrule
     T2 lesion load &- &SVM & \textbf{88.46 \%} & 76.92 \% &  \textbf{100.00} \% & 94.62 \% \\
     FLAIR & - & SVM & 66.92 \% & 53.85 \% & 80.00 \% & 66.92 \% \\
     \midrule
     FLAIR & no & CNN &  71.23 \% & 68.46 \% & 74.00 \% & 85.46 \% \\
     FLAIR & yes & CNN &   87.04 \% &  \textbf{93.08} \% &  81.00 \%  & \textbf{96.08} \% \\
     \midrule
     FLAIR - les. fill. & yes  & CNN &   70.15 \% & 92.31 \% &  48.00 \% & 90.92 \%  \\
     \bottomrule
   \end{tabular}
   \caption{Performance (in \%) for the different models on the holdout data set. Pre-train., pre-training; Class., classifier; Bal. acc., balanced accuracy; Sens., sensitivity; Spec., specificity; AUC, area under the curve of the receiver operating characteristic; les. fill., lesions filled.}
   \label{res-table}
 \end{table*}

\subsection{Visualization}
  
\begin{figure*}
    \centering
    \includegraphics[width=400pt]{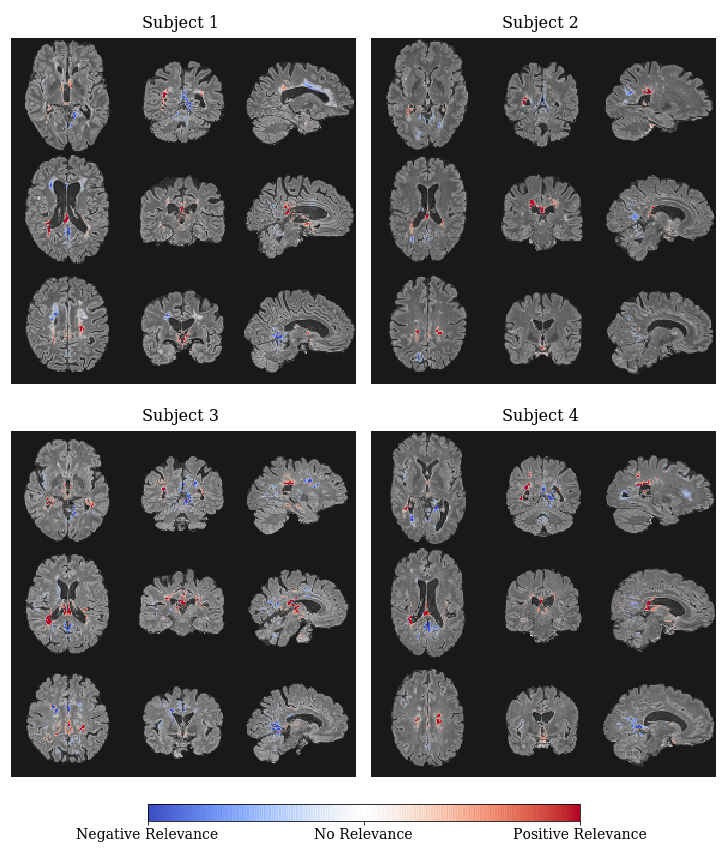}
    \caption{Individual LRP heatmaps (overlayed on the input FLAIR data) for the four MS patients with the highest classification score. Heatmap values are normalized in the range [-0.03, 0.03]. Colors indicate regions supporting (red) or rejecting (blue) the classification as a MS patient with respect to the underlying CNN model.}
    \label{fig:individuals_comp}
\end{figure*}

After the CNN models have been trained, we used LRP to generate an individual heatmap for each subject in the holdout data set indicating the relevance of each voxel for the respective classification decision. Unless otherwise stated, we restricted our analyses only to correctly classified examples (i.e. true positives and true negatives). In Figure \ref{fig:individuals_comp}, we show the individual heatmaps overlayed on the FLAIR data for four MS patients, who achieved the highest classification scores. High classification scores generally indicate a higher confidence of the model for the respective classification decision and thus the corresponding explanations are usually more pronounced and less diffuse as for cases with lower classification scores. 
All four patients have in common that high positive relevance is attributed around the occipital horn of both lateral ventricles and covers periventricular lesion areas as well as the body and splenium of the corpus callosum. Even though the images were clearly classified as MS, certain regions are assigned negative relevance, meaning that these areas speak against the MS diagnosis. Negative relevance can be found around the frontal horn of both ventricles, notably even in periventricular lesion areas (see for example subject 1). Interestingly, lesions not bordering the ventricles seem often to be ignored or are assigned negative relevance. 
Please note that -- in contrast to classical multi-univarate studies -- the CNN model in combination with LRP is invariant to translations and thus is capable of identifying specific tissue areas (e.g. white matter lesions) having a positive relevance for MS regardless of their position in the voxel space.

\begin{figure}[H]
    \centering
    \includegraphics[width=\linewidth]{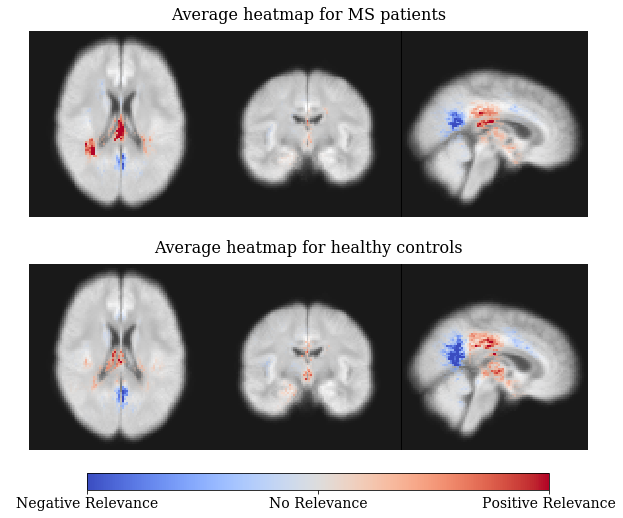}
    \caption{Average LRP heatmaps for all correctly classified MS patients (top) and all correctly classified healthy controls (bottom) in the holdout set. Values are normalized in the range [-0.02, 0.02].}
    \label{fig:relevance_average_TP_TN}
\end{figure}

In Figure \ref{fig:relevance_average_TP_TN}, we show average heatmaps for all correctly classified MS patients (top) and all correctly classified healthy controls (bottom) in the holdout set. In accordance with the heatmaps of the individual subjects in Figure \ref{fig:individuals_comp}, posterior periventricular white matter regions have a strong positive relevance for the MS diagnosis. This is true for both MS patients and healthy controls, but the effect is less pronounced for healthy controls. The reversed effect can be seen for clusters exhibiting negative relevance in white matter areas in the corpus callosum and close to occipital and parietal lobe. Over all voxels healthy controls typically obtain a negative relevance sum (mean$\pm$std: -1.05e-6 $\pm$ 0.0013) as opposed to a positive relevance sum in MS patients (3.07e-06 $\pm$ 0.0014).

\begin{figure*}
    \centering
    \subfloat[]{\includegraphics[width=0.5\linewidth]{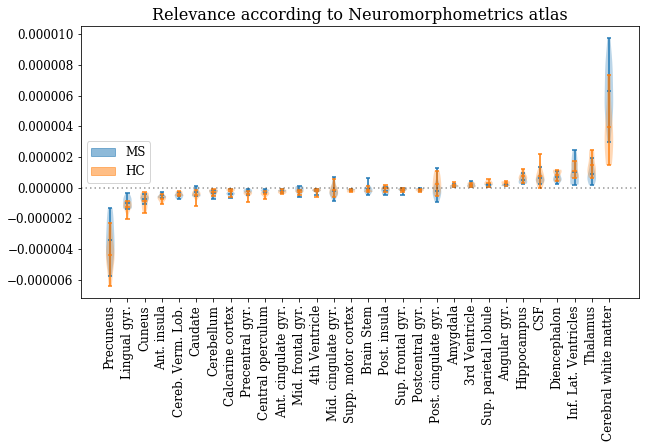}\label{fig:relevance_GM}}
    \subfloat[]{\includegraphics[width=0.5\linewidth]{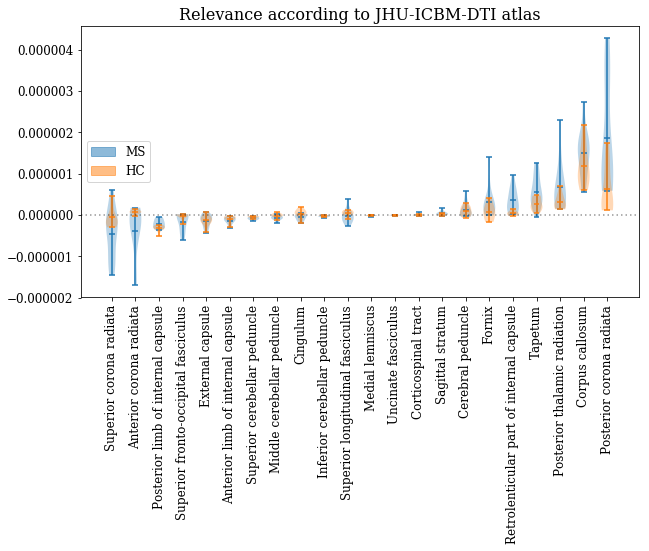}\label{fig:relevance_WM}}
    \caption{LRP relevance distribution over (a) 30 (mainly) gray matter areas from the Neuromorphometrics atlas and (b) 22 white matter areas from the JHU ICBM-DTI atlas, separately for MS patients and healthy controls in the holdout set.}
    \label{fig:relevance_both}
\end{figure*}

To analyze the LRP heatmaps quantitatively with respect to different brain areas, we computed the relevance sum in (1) the Neuromorphometrics atlas mostly containing gray matter regions and (2) the JHU ICBM-DTI atlas containing white matter regions. Areas were aggregated between left and right hemisphere and certain substructures are combined into one region. For visualization of (1) we selected the 30 areas with the highest sum of absolute relevance means across MS patients and healthy controls in the test set, yielding areas with both the highest and lowest relevance. Please reconsider here that the MRI data have only been linearly registered and thus slight deviations from the anatomical locations stated in the atlases are conceivable. 
 In Figure \ref{fig:relevance_both}, we depict the region-wise LRP relevance for MS diagnosis, separately for MS patients and healthy controls. In the Neuromorphometrics atlas (see Figure \ref{fig:relevance_GM}), most relevance is attributed to cerebral white matter, followed by thalamus, lateral ventricles and diencephalon. 
Negative relevance is strongest in the precuneus, followed by lingual gyrus, cuneus and insula. In the JHU white matter atlas (see Figure \ref{fig:relevance_WM}), most positive relevance is attributed to posterior corona radiata and corpus callosum, followed by posterior thalamic radiation, tapetum, internal capsule and fornix. Notably, these areas are generally characterized by a high lesion density, which is also present in this MS data set (see supplementary Figures 2 and 3). Negative relevance has been found in the superior and anterior corona radiata. Generally, the relevance for MS patients is higher in white matter than in gray matter areas. Moreoever, the differences between MS patients and healthy controls are more pronounced in white matter areas.

\captionsetup*[table]{position=top}
\captionsetup*[subtable]{position=top}
\captionsetup*[figure]{position=bottom}
\captionsetup[subfigure]{position=top, labelfont=bf,textfont=normalfont,singlelinecheck=off,justification=raggedright}

\begin{figure*}
    \centering
    \includegraphics[width=300pt]{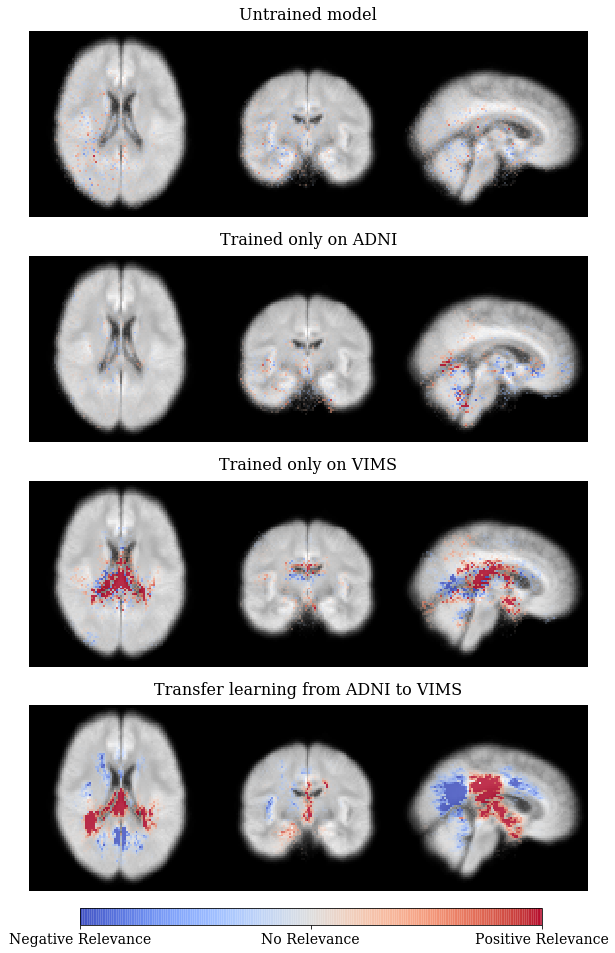}
    \caption{Average heatmaps for different CNN models applied to the MS (VIMS) cohort -- starting from an untrained CNN model with random parameters over a CNN trained only on either ADNI or MS data to a CNN pre-trained on ADNI and fine-tuned on MS. As it can be seen, the fine-tuned model led to the most concise regions of positive and negative relevance. Please note that we averaged here the heatmaps over all (not only the correctly classified) MS patients in the holdout set and that the heatmap values here are not normalized to a fixed range but shown with respect to the minimum value of the untrained model. 
    }
    \label{fig:relevance_transfer}
\end{figure*}

In Figure \ref{fig:relevance_transfer}, we show the effects of transfer learning on the average relevance heatmaps for the MS patients in the holdout set. For the untrained model with random parameters (first row), only scarcely distributed individual voxels attain tiny relevance values. For the CNN model trained on ADNI and directly applied to MS patients (without fine-tuning; second row), more voxels are attributed relevance and are diffusely clustered. For the CNN model trained only on MS data (without pre-training; third row), strong relevance is projected to the ventricles and periventricular white matter. And finally, for the pre-trained model (transfer learning from ADNI to MS; last row), distinct clusters for both positive and negative relevance can be detected, which are more delineated than for the CNN model without pre-training.

\begin{figure}
    \centering
    \includegraphics[width=\linewidth]{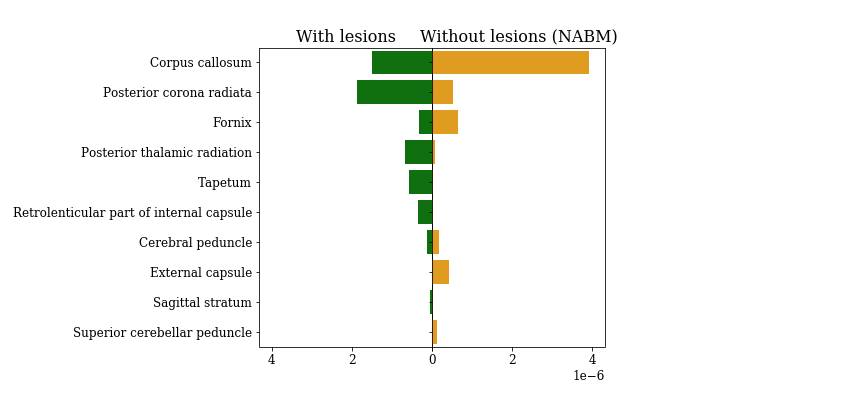}
    \caption{Comparison of average relevance distribution over white matter areas for a CNN model trained on original FLAIR data (left) and lesion-filled FLAIR data (right; NABM, normal-appearing brain matter). We calculated the relevance sum of both models (averaged over subjects) and show the 10 areas with the highest score.}
    \label{fig:lesion_removal_comparison}
\end{figure}

To assess the contribution of normal-appearing brain matter, we compared the relevance maps between the CNN models trained on the original FLAIR data and the lesion-filled FLAIR data (for the performance see Table \ref{res-table}). In Figure \ref{fig:lesion_removal_comparison}, we depict the relevance for the 10 top-scored white matter regions, separately for both models. 
In general one can see that the relevance shifts from a distribution more evenly spread among multiple areas to a distribution with a prominent peak and otherwise low shares of relevance. Notably, relevance is shifted away from areas with large amounts of lesions such as posterior corona radiata, posterior thalamic radiata as well as tapetum towards mainly the corpus callosum and regions with very few lesions like fornix and external capsule (see supplementary Figure 2 for distribution of white matter lesions).

\section{Discussion}
\label{S:discussion}

\subsection{Summary}
In the present study, we introduced a transparent framework for analyzing neuroimaging data with CNNs that is able to explain individual classification decisions. By utilizing transfer learning we could further achieve good classification results from only a small data set of task-specific data. In combination with layer-wise relevance propagation (LRP), we could demonstrate the capacity of our framework to learn significant MS-relevant information from conventional MRI data. Notably, a pre-trained CNN was able to identify MS patients with an accuracy similar to a classical machine learning analysis, in which the T2 lesion load was used as input. This is quite remarkable, because the CNN model was considered to be naive by not being provided with any prior information on MS-relevant features such as hyperintense lesions. 
The subsequent visualization analysis, using heatmaps generated by LRP, revealed that the CNN model indeed uses (posterior) white matter lesions as primary information source. In addition, other information, e.g. in normal-appearing white and gray matter (e.g. the thalamus) have been found useful by the CNN model.

\subsection{Key findings}

\textbf{CNNs learn to identify lesions as an important biomarker for MS.}
Although our pre-trained CNN model did not get any prior information about the relevance of hyperintense lesions for MS, it learned to successfully identify lesions as a primary information source. Notably, 9.71\% of total relevance was attributed to lesion areas compared to a lesion coverage of only  0.44\% in the training data set. We show that LRP heatmaps not only detect single lesions in individual patients but generally attributed most positive relevance to white matter areas around the posterior occipital horns. Importantly, the CNN model did not simply assign high relevance to hyperintense areas in the brain, but learned to distinguish between different lesion locations: while anterior periventricular lesions as well as lesions not bordering the lateral ventricles were assigned no or negative relevance, only posterior periventricular lesion areas were assigned positive relevance for MS. Strongest positive relevance was found for posterior corona radiata, corpus callosum and thalamic radiation, which are generally characterized by a high lesion density in MS patients (see \citep{Filli2012} and supplementary Figures 2 and 3). 

\textbf{CNNs learn to identify relevant areas beyond lesions.}
The CNN model primarily focuses on lesions, but relevance has also been attributed to gray matter areas such as the thalamus, which is known to be affected in MS \citep{Azevedo2018}. To further investigate what the CNN model learns beyond lesions, we repeated the analysis on lesion filled MRI data. As expected, the balanced accuracy as well as AUC decreased (by almost 17 and 6 percentage points respectively) and relevance has shifted away from regions which typically contain hyperintense lesions. 
The region that was assigned most relevance after lesion removal was the corpus callosum. While the corpus callosum is generally susceptible to demyelinating lesions \citep{Barnard1974, Garg2015, Renard1041} the literature also suggests further biomarkers such as axonal loss and diffuse atrophy \citep{Renard1041,Evangelou2000} or narrow T2 hyperintense bands along the callosal-septal interface \citep{Garg2015}. The fornix, even though it contains a very small amount of lesions (see supplementary Figure 2 and \citep{thomas2011Fornix}), is assigned positive relevance with lesions and an increased relevance without lesions. It has been shown that lower fractional anisotropy in the fornix is exhibited in MS subjects in comparison to healthy controls \citep{Roosendaal2009, Kern2012}. Additionally, external capsule and superior cerebellar peduncle receive only positive relevance after lesion removal, which were found to be affected in MS patients \citep{Anderson2011cerebellar, han2017capsule}. 
These results are generally in line with other machine learning studies finding differences in normal-appearing brain matter in MS patients \citep{Weygandt2011, Hackmack2012a, Yoo2018}. 
It would be very interesting to further investigate whether our findings correlate with underlying pathological mechanisms only demonstrable by advanced MRI sequences such as diffusion weighted imaging or magnetization transfer imaging.

\textbf{Transfer learning improves learning on small clinical neuroimaging cohorts, even across diseases and MRI sequences.}
In recent years, transfer learning has been successfully employed in brain lesion segmentation \citep{Ghafoorian2017TransferLearning} and Alzheimer's disease classification \citep{Gupta2013, Payan2015, Hosseini-Asl2018}. The latter studies used either autoencoders trained on MRI data or natural images \citep{Gupta2013, Payan2015} or used one Alzheimer's disease data set for pre-training and another Alzheimer's disease data set for fine-tuning \citep{Hosseini-Asl2018}. In the present study, we have shown that transfer learning can also help in learning (1) across diseases (Alzheimer's disease to MS) and (2) across MRI sequences (MPRAGE to FLAIR). We demonstrated that not only the balanced accuracy increases drastically (about 16 percentage points), but also that LRP leads to much more focused heatmaps concentrating on (posterior) periventricular lesion areas.  
Given that our pre-trained model performed similar to a classical machine learning analysis using T2 lesion load as a classical biomarker in MS, we believe that larger data sets might allow for outperforming models based on lesion masks in the future. Additionally, we are convinced that our approach -- given a reasonable data basis --  might also be very useful in answering more complex questions such as predicting disease progression.

\subsection{Related work}
Compared to other neurological diseases, in particular Alzheimer's disease, only a few MS studies exist that employ machine learning methods outside the scope of lesion segmentation. We think that the main reasons are (1) the lack of easy accessible large open data bases such as the Alzheimer's Neuroimaging Initiative (ADNI) data base and (2) the focus on white matter lesion volume as primary MRI-derived outcome measure in MS. Classical machine learning methods in combination with more or less sophisticated feature extraction methods, from both conventional and advanced MRI data, have been used to (1) diagnose MS \citep{Weygandt2011,Hackmack2012b, ZURITA2018724} (2) decode symptom severity \citep{Hackmack2012a} (3) identify clinical subtypes \citep{Eshaghi2015, Eshaghi2018, Bendfeldt2012} and (4) predict conversion from clinically isolated syndrome to MS \citep{Wottschel2015}.  
Deep learning architectures have so far been implemented for lesion segmentation  \citep{Valverde2017, Brosch2016, Khastavaneh2017}, predicting MS based on binary lesion masks \citep{Yoo2016b}, modelling brain and lesion variability \citep{Brosch2016a} and finding differences in normal-appearing brain matter based on T1-weighted and myelin images \citep{Yoo2018}. To the best of our best knowledge, the present study is the first study employing CNNs and advanced visualization techniques for diagnosing MS based on FLAIR data. 

It is generally recognized that, especially in the medical field, it is very important that classification decisions are reasonably explained even in light of high accuracies (which are no guarantee for a -- from a human perspective -- sensible discrimination strategy \citep{Lapuschkin2016,Lapuschkin2019}).
Although a number of methods exist that generate individual heatmaps \citep{Simonyan2013, Springenberg2015, Zeiler2014, Zintgraf2017}, we focused here on the LRP method \citep{Bach2015, Montavon2018,Lapuschkin2019} which has a solid theoretical framework and has been extensively validated (see e.g. \citep{Samek2015, samek2017explainable, Lapuschkin2019}). Very recently, LRP has shown to be very helpful for explaining cognitive states or Alzheimer's disease diagnosis in deep neural networks trained on either functional or structural MRI data \citep{boehle2019Visualizing,Thomas2018}. In the present study, we demonstrated that LRP is capable of identifying reasonable areas supporting a MS diagnosis in addition to features needing further clinical validation. 
By this, we have shown that those heatmaps can be very valuable in explaining decisions of neural networks trained on small sample sizes and to verify whether an algorithm has learned something meaningful (i.e. matching domain knowledge) or just spotted biases or artifacts in the data (see also \citep{Bach2015, Lapuschkin2019}).

 \subsection{Limitations}
The main limitation of this study is the limited sample size. Although a sample size of $n=147$ is comparable with other deep learning studies in the neuroimaging field \citep{Vieira2017}, it is generally considered to be too low to learn robust representations from the data and to generalize to other data sets. To partly alleviate this problem, we pre-trained our network on ADNI data ($n=921$) and fine-tuned it on the MS data. By visualizing the average heatmaps for MS patients, we show in addition to a balanced accuracy of $87.04$ \% that the CNN captures MS-relevant information by focusing on posterior ventricular regions usually characterized by a high rate of MS lesion incidences. Nevertheless, future studies should verify our results in larger data sets, preferably coming from different sites. Another limitation, related to the first one, is that we were limited in the choice of architecture used for the CNN analysis. To avoid overfitting, we have chosen a relatively simple CNN architecture and included different methods for regularization (drop out, L2-regularization and early stopping). Moreover, by registering the MRI data only linearly to MNI space, the regions contained in both atlases only roughly correspond to individual anatomical locations. On the other hand, non-linear registration can lead to strong deformations, in particular in patients, and we show here that our CNN model can also operate on a more native level (in accordance with \citep{Suk2014}). To be able to make more specific anatomical claims in individual subjects, future studies might use individual atlases. And finally, heatmaps do neither allow to determine the underlying pathological mechanism (e.g. atrophy, demyelination or axonal loss) resulting in assigning a voxel to be relevant or to assess interactions between voxels. For this, one would have to take a deeper look into the specific filters that have been learned throughout the training process in combination with MR sequences more sensitive for certain tissue damage (e.g. diffusion weighted imaging). Nevertheless, we still believe that heatmaps can be very helpful in supplementing individual disease diagnoses by providing a simple and intuitive explanation.

\section{Conclusion}
\label{S:conclusion}
In conclusion, we have shown that CNN models are capable of learning MS-relevant information from a typical-sized neuroimaging data set. In particular, we demonstrated that pre-training on additional data substantially increases the prediction performance (even across diseases and MRI sequences) and that the LRP method is very valuable not only in explaining individual network's decisions, but also in generally helping to assess whether CNN models have learned significant features. Notably, our CNN models focus on hyperintense lesions as primary information source, but also incorporates information from lesion location and normal-appearing brain areas. We see a high potential in the combination of CNNs, transfer learning and LRP heatmaps and are convinced that our framework might not only be helpful in other disease decoding studies, but also for answering more complex questions such as predicting disease progression or treatment response in individual subjects. 
\section{Funding}
We acknowledge support from the German Research Foundation (DFG, 389563835), the Manfred and Ursula-M\"uller Stiftung and Charit\'e -- Universit\"atsmedizin Berlin (Rahel-Hirsch scholarship).


 \section{References}



\small
\bibliographystyle{model1-num-names}
 \bibliography{references.bib}








\end{document}


\section{Supplementary Materials}



\begin{figure}[H]
    \centering
    \includegraphics[width=250pt]{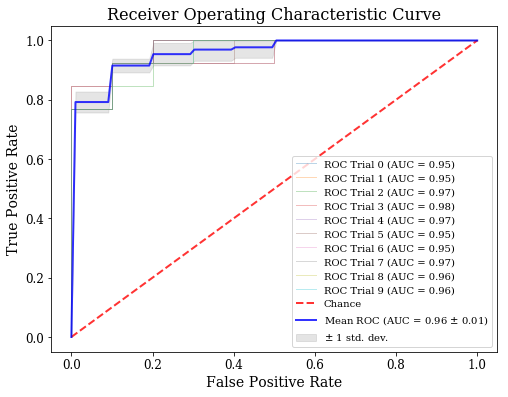}
    \caption{Performance of the pre-trained CNN model (fine-tuned on MS) on the holdout set.}
    \label{fig:ROC}
\end{figure}

\begin{figure}[H]
    \centering
    \includegraphics[width=310pt]{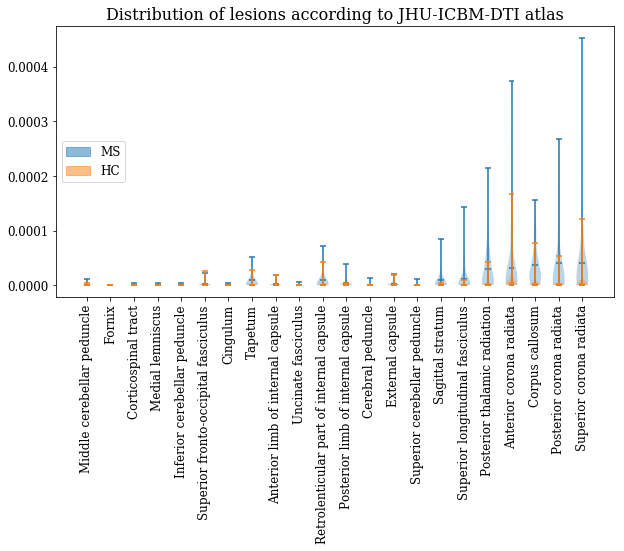}
    \caption{Lesion distribution over white matter areas from the JHU ICBM-DTI atlas, separately for MS patients and healthy controls (HC).}
    \label{fig:lesion_distribution_WM}
\end{figure}

\begin{figure}[H]
    \centering
    \includegraphics[width=310pt]{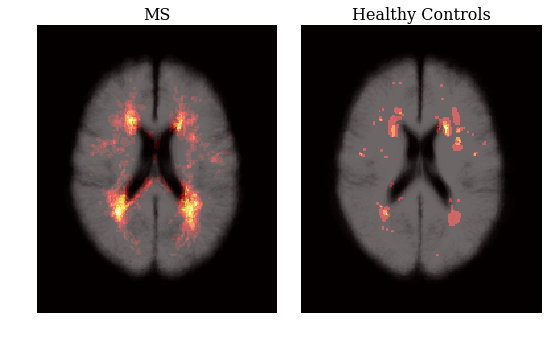}
    \caption{Lesion distribution in MS patients (left) and healthy controls (right).}
    \label{fig:lesion_distribution_on_average}
\end{figure}